\documentstyle[aclap]{article}

\newcommand{\caplab}[2]{
        \caption{\label{#1}#2}}

\newcommand{\figbegin}{
        \begin{figure}[htbp]
        \begin{quote}
        \small}

\newcommand{\figend}[2]{
        \end{quote}
        \caplab{#1}{#2}
        \end{figure}}

\newcommand{\inspsfig}[9]{
        \figbegin{
                \begin{picture}(#3,#4)
                \put(#5,#6){\includegraphics{#7}}
                \end{picture}}
        \figend{#1}{#2}}

\def\argmax{\mathop{\rm argmax}}

\title{\vspace{-0.5in}Comparing a Linguistic and a Stochastic Tagger}
\author{Christer Samuelsson \\
 Lucent Technologies\\
 Bell Laboratories\\
 600 Mountain Ave, Room 2D-339\\
 Murray Hill, NJ 07974, USA\\
 {\tt christer@research.bell-labs.com}
\And
 Atro Voutilainen\\
 Research Unit for Multilingual Language Technology\\
 P.O. Box 4\\
 FIN-00014 University of Helsinki\\
 Finland\\
 {\tt Atro.Voutilainen@Helsinki.FI}
       }

\begin{document}
\bibliographystyle{fullname}
\maketitle
\vspace{-0.5in}
\begin{abstract}
  Concerning different approaches to automatic PoS tagging:
  EngCG-2, a constraint-based morphological tagger,
  is compared in a double-blind test with a state-of-the-art
  statistical tagger on a common disambiguation task using a common
  tag set.  The experiments show that for the same amount of remaining
  ambiguity, the error rate of the statistical tagger is one order of
  magnitude greater than that of the rule-based one.  The two related
  issues of priming effects compromising the results and disagreement
  between human annotators are also addressed.
\end{abstract}

\section{Introduction\footnote{Published in {\it Proceedings of 35th Annual
Meeting of the Association for Computational Linguistics and 8th
Conference of the European Chapter of the Association for
Computational Linguistics}. ACL, Madrid.}}

There are currently two main methods for automatic part-of-speech
tagging.  The prevailing one uses essentially statistical language
models automatically derived from usually hand-annotated corpora.
These corpus-based models can be represented e.g.\ as collocational
matrices (Garside et al.\ (eds.) 1987; Church 1988), Hidden Markov
models (cf.\ Cutting et al.\ 1992), local rules (e.g.\ Hindle 1989)
and neural networks (e.g.\ Schmid 1994).  Taggers using these
statistical language models are generally reported to assign the
correct and unique tag to 95-97\% of words in running text, using tag
sets ranging from some dozens to about 130 tags.

The less popular approach is based on hand-coded linguistic
rules.  Pioneering work was done in the 1960's (e.g.\ Greene and Rubin
1971).
Recently, new interest in the
linguistic approach has been shown e.g.\ in the work of (Karlsson
1990; Voutilainen et al.\ 1992; Oflazer and Kuru{\"o}z 1994; Chanod and
Tapanainen 1995; Karlsson et al.\ (eds.) 1995; Voutilainen 1995). The
first serious linguistic competitor to data-driven statistical taggers
is the English Constraint Grammar parser, EngCG (cf.\ Voutilainen et
al.\ 1992; Karlsson et al.\ (eds.) 1995).  The tagger consists of the
following sequentially applied modules:

\begin{enumerate}
    \item Tokenisation
    \item Morphological analysis
    \begin{enumerate}
      \item    Lexical component
      \item    Rule-based guesser for unknown words
    \end{enumerate}
    \item Resolution of morphological ambiguities
\end{enumerate}

The tagger uses a two-level morphological analyser with a large lexicon and
a morphological description that introduces about 180 different
ambiguity-forming morphological analyses, as a result of which each
word gets 1.7-2.2 different analyses on an average. Morphological
analyses are assigned to unknown words with an accurate rule-based
`guesser'. The morphological disambiguator uses constraint rules that
discard illegitimate morphological analyses on the basis of local or
global context conditions. The rules can be grouped as ordered
subgrammars: e.g.\ heuristic subgrammar 2 can be applied for resolving
ambiguities left pending by the more `careful' subgrammar 1.

Older versions of EngCG (using about 1,150 constraints) are reported
(Voutilainen et al.\ 1992; Voutilainen and Heikkil{\"a} 1994; Tapanainen
and Voutilainen 1994; Voutilainen 1995) to assign a correct analysis
to about 99.7\% of all words while each word in the output retains
1.04-1.09 alternative analyses on an average, i.e.\ some of the
ambiguities remain unresolved.

These results have been seriously questioned.  One doubt concerns the
notion ``correct analysis''. For example Church (1992) argues that
linguists who manually perform the tagging task using the double-blind
method disagree about the correct analysis in at least 3\% of all
words even after they have negotiated about the initial disagreements.
If this were the case, reporting accuracies above this 97\% `upper
bound' would make no sense.

However, Voutilainen and J{\"a}rvinen (1995) empirically show that an
interjudge agreement virtually of 100\% is possible, at least
with the EngCG tag set if not with the original Brown Corpus tag set.
This consistent applicability of the EngCG tag set is explained by
characterising it as grammatically rather than semantically
motivated.

Another main reservation about the EngCG figures is the suspicion
that, perhaps partly due to the somewhat underspecific nature of the
EngCG tag set, it must be so easy to disambiguate that also a
statistical tagger using the EngCG tags would reach at least as good
results. This argument will be examined in this paper. It will be
empirically shown (i) that the EngCG tag set is about as difficult for
a probabilistic tagger as more generally used tag sets and (ii) that
the EngCG disambiguator has a clearly smaller error rate than the
probabilistic tagger when a similar (small) amount of ambiguity is
permitted in the output.

A state-of-the-art statistical tagger is trained on a corpus of over
350,000 words hand-annotated with EngCG tags, then both taggers (a new
version known as EngCG-2\footnote{An online version of EngCG-2 can be
  found at\\ 
  http:\-//www.\-ling.\-helsinki.\-fi/\~{}avoutila/\-engcg-2.html.}
with 3,600 constraints as five subgrammars\footnote{The first three
  subgrammars are generally highly reliable and almost all of the
  total grammar development time was spent on them; the last two
  contain rather rough heuristic constraints.}, and a statistical
tagger) are applied to the same held-out benchmark corpus of 55,000
words, and their performances are compared.  The results disconfirm
the suspected `easiness' of the EngCG tag set: the statistical
tagger's performance figures are no better than is the case with
better known tag sets.

Two caveats are in order.  What we are not addressing in this paper is
the work load required for making a rule-based or a data-driven
tagger. The rules in EngCG certainly took a considerable effort to
write, and though at the present state of knowledge rules could be
written and tested with less effort, it may well be the case that a
tagger with an accuracy of 95-97\% can be produced with less effort by
using data-driven techniques.\footnote{However, for an interesting
  experiment suggesting otherwise, see \cite{CT:95}.}

Another caveat is that EngCG alone does not resolve all ambiguities,
so it cannot be compared to a typical statistical tagger if full
disambiguation is required. However, Voutilainen (1995) has shown that
EngCG combined with a syntactic parser produces morphologically
unambiguous output with an accuracy of 99.3\%, a figure clearly better
than that of the statistical tagger in the experiments below (however,
the test data was not the same).

Before examining the statistical tagger, two practical points are
addressed: the annotation of the corpora used, and the modification of
the EngCG tag set for use in a statistical tagger.

\section{Preparation of Corpus Resources}

\subsection{Annotation of training corpus}

The stochastic tagger was trained on a sample of 357,000 words from
the Brown University Corpus of Present-Day English \cite{Francis:82}
that was annotated using the EngCG tags. The corpus was first analysed
with the EngCG lexical analyser, and then it was fully disambiguated
and, when necessary, corrected by a human expert. This annotation took
place a few years ago. Since then, it has been used in the development
of new EngCG constraints (the present version, EngCG-2, contains about
3,600 constraints): new constraints were applied to the training
corpus, and whenever a reading marked as correct was discarded, either
the analysis in the corpus, or the constraint itself, was corrected.
In this way, the tagging quality of the corpus was continuously
improved.

\subsection{Annotation of benchmark corpus}

Our comparisons use a held-out benchmark corpus of about 55,000 words
of journalistic, scientific and manual texts, i.e., no training
effects are expected for either system. The benchmark corpus was
annotated by first applying the preprocessor and morphological
analyser, but not the morphological disambiguator, to the text.
This morphologically ambiguous text was then
independently and fully disambiguated by two experts whose task was
also to detect any errors potentially produced by the previously
applied components. They worked independently, consulting written
documentation of the tag set when necessary. Then these manually
disambiguated versions were automatically compared with each other. At
this stage, about 99.3\% of all analyses were identical. When the
differences were collectively examined, virtually all were agreed to
be due to clerical mistakes.  Only in the analysis of 21 words,
different (meaning-level) interpretations persisted, and even here
both judges agreed the ambiguity to be genuine. One of these two
corpus versions was modified to represent the consensus, and this
`consensus corpus' was used as a benchmark in the evaluations.

As explained in Voutilainen and J{\"a}rvinen (1995), this high agreement
rate is due to two main factors. Firstly, distinctions based on some
kind of vague semantics are avoided, which is not always case with
better known tag sets. 
Secondly, the adopted analysis of most of the constructions where
humans tend to be uncertain is documented as a collection of tag
application principles in the form of a grammarian's manual (for
further details, cf.\ Voutilainen and J{\"a}rvinen 1995).

The corpus-annotation procedure allows us to perform a
text-book statistical hypothesis test. Let the null hypothesis be that
any two human evaluators will necessarily disagree in at least 3\% of
the cases.  Under this assumption, the probability of an observed
disagreement of less than 2.88\% is less than 5\%.  This can be
seen as follows: For the relative frequency of disagreement, $f_n$, we
have that $f_n$ is approximately $\sim N(p,\sqrt{\frac{p(1-p)}{n}})$,
where $p$ is the actual disagreement probability and $n$ is the number
of trials, i.e., the corpus size.  This means that
$P((\displaystyle\frac{f_n-p}{\sqrt{p(1-p)}}\sqrt{n} \le x) \approx \Phi(x)$
where $\Phi$ is the standard normal distribution function.  This in turn
means that
\begin{eqnarray*}
  P(f_n \le p + x \sqrt{\frac{p(1-p)}{n}}) &\approx& \Phi(x)
\end{eqnarray*}
Here $n$ is 55,000 and $\Phi(-1.645) = 0.05$.  Under the null
hypothesis, $p$ is at least 3\% and thus:
\begin{eqnarray*}
  \lefteqn{ P(f_n \le 0.03 - 1.645 \sqrt{\frac{0.03 \cdot
        0.97}{55,000}}) \;\;=}\\ &=& P(f_n \le 0.0288) \;\;\le\;\;
  0.05 \;\;\;\;\;\;\;\;\;\;\;\;
\end{eqnarray*}
We can thus discard the null hypothesis at significance level 5\%
if the observed disagreement is less than 2.88\%.  It was in
fact 0.7\% before error correction, and virtually zero
($\rule{0mm}{6mm}\displaystyle\frac{21}{55,000}$) after negotiation.
This means that we can
actually discard the hypotheses that the human evaluators in average
disagree in at least 0.8\% of the cases before error correction, and
in at least 0.1\% of the cases after negotiations, at significance
level 5\%.

\subsection{Tag set conversion}

The EngCG morphological analyser's output formally differs from most
tagged corpora; consider the following 5-ways ambiguous analysis of
``walk'':

\begin{verbatim}
walk
   walk <SV> <SVO> V SUBJUNCTIVE VFIN
   walk <SV> <SVO> V IMP VFIN
   walk <SV> <SVO> V INF
   walk <SV> <SVO> V PRES -SG3 VFIN
   walk N NOM SG
\end{verbatim}

Statistical taggers usually employ single tags to indicate analyses
(e.g.\ ``NN'' for ``N NOM SG''). Therefore a simple conversion program
was made for producing the following kind of output, where each
reading is represented as a single tag:

\begin{verbatim}
walk V-SUBJUNCTIVE V-IMP V-INF
     V-PRES-BASE N-NOM-SG
\end{verbatim}

The conversion program reduces the multipart EngCG tags into a set of
80 word tags and 17 punctuation tags (see Appendix) that retain the
central linguistic characteristics of the original EngCG tag set.

A reduced version of the benchmark corpus was prepared with this
conversion program for the statistical tagger's use. Also EngCG's
output was converted into this format to enable direct comparison with
the statistical tagger.

\section{The Statistical Tagger}

The statistical tagger used in the experiments is a classical
trigram-based HMM decoder of the kind described in e.g.\ 
\cite{Church:88}, \cite{DeRose:88} and numerous other articles.
Following conventional notation, e.g. \cite[pp.\ 272-274]{Rabiner:89} and
\cite[pp.\ 42-46]{K-S:96}, the tagger recursively calculates the
$\alpha$, $\beta$, $\gamma$ and $\delta$ variables for each word
string position $t=1,\ldots,T$ and each possible state%
\footnote{The {\it N-1\/}th-order HMM corresponding to an N-gram tagger
is encoded as a first-order HMM, where each state corresponds to
a sequence of {\it N-1\/} tags, i.e., for a trigram tagger, each state
corresponds to a tag pair.}
$s_i : i=1,\ldots,n$:
\begin{eqnarray*}
\alpha_t(i) &=& P(\mbox{\bf W}_{\le t};S_t=s_i)\\
\beta_t(i) &=& P(\mbox{\bf W}_{> t} \mid S_t=s_i)\\
\gamma_t(i) &=& P(S_t=s_i \mid \mbox{\bf W}) \;\;=\;\;
\frac{P(\mbox{\bf W};S_t=s_i)}{P(\mbox{\bf W})} \;\;=\\
&=& \frac{\alpha_t(i) \cdot \beta_t(i)}
  {\displaystyle \sum_{i=1}^n \alpha_t(i) \cdot \beta_t(i)}\\ 
\delta_t(i) &=& 
\max_{\mbox{\bf S}_{\le t-1}}
P(\mbox{\bf S}_{\le t-1},S_t=s_i;\mbox{\bf W}_{\le t})
\end{eqnarray*}
Here
\begin{eqnarray*}
  \mbox{\bf W} &=& W_1=w_{k_1},\ldots,W_T=w_{k_T}\\ \mbox{\bf W}_{\le
    t} &=& W_1=w_{k_1},\ldots,W_t=w_{k_t}\\ \mbox{\bf W}_{> t} &=&
  W_{t+1}=w_{k_{t+1}},\ldots,W_T=w_{k_T}\\ \mbox{\bf S}_{\le t} &=&
  S_1=s_{i_1},\ldots,S_t=s_{i_t}
\end{eqnarray*}
where $S_t=s_i$ is the event of the $t$th word being emitted
from state $s_i$ and $W_t=w_{k_t}$ is the event of the $t$th word
being the particular word $w_{k_t}$ that was actually observed in the
word string.

Note that for $t=1,\ldots,T-1$ ; $i,j=1,\ldots,n$
\begin{eqnarray*}
\alpha_{t+1}(j) &=& 
\left[\sum_{i=1}^n \alpha_t(i) \cdot p_{ij}\right] \cdot a_{jk_{t+1}}\\
\beta_t(i) &=& \sum_{j=1}^n \beta_{t+1}(j) \cdot p_{ij} \cdot a_{jk_{t+1}}\\
\delta_{t+1}(j) &=&
\left[\max_i \delta_t(i) \cdot p_{ij}\right] \cdot a_{jk_{t+1}}
\end{eqnarray*}
where $p_{ij} = P(S_{t+1}=s_j \mid S_t=s_i)$
are the transition probabilities, encoding the tag N-gram probabilities, and
\begin{eqnarray*}
\lefteqn{a_{jk} \;\;=}\\
&=& P(W_t=w_k \mid S_t=s_j) \;=\; P(W_t=w_k \mid X_t=x_j)
\end{eqnarray*}
are the lexical probabilities.
Here $X_t$ is the random variable of assigning a tag to the $t$th word
and $x_j$ is the last tag of the tag sequence encoded as state $s_j$.
Note that $s_i \ne s_j$ need not imply $x_i \ne x_j$.

More precisely, the tagger employs the converse lexical probabilities
\begin{eqnarray*}
a^\prime_{jk} &=& \frac{P(X_t=x_j \mid W_t=w_k)}{P(X_t=x_j)} \;\;=\;\;
\frac{a_{jk}}{P(W_t=w_k)}
\end{eqnarray*}
This results in slight variants $\alpha^\prime$, $\beta^\prime$,
$\gamma^\prime$ and $\delta^\prime$ of the original quantities:
\begin{eqnarray*}
  \frac{\alpha_t(i)}{\alpha_t^\prime(i)} &=&
  \frac{\delta_t(i)}{\delta_t^\prime(i)} \;\;=\;\; \prod_{u=1}^t
  P(W_u=w_{k_u})\\
\frac{\beta_t(i)}{\beta_t^\prime(i)} &=& \prod_{u=t+1}^T P(W_u=w_{k_u})
\end{eqnarray*}
and thus $\forall i,t$
\begin{eqnarray*}
\gamma_t^\prime(i) &=& 
\frac{\alpha_t^\prime(i) \cdot \beta_t^\prime(i)}
     {\displaystyle \sum_{i=1}^n \alpha_t^\prime(i) \cdot \beta_t^\prime(i)}
\;\;=\\
&=& \frac{\alpha_t(i) \cdot \beta_t(i)}
         {\displaystyle \sum_{i=1}^n \alpha_t(i) \cdot \beta_t(i)}
\;\;=\;\; \gamma_t(i)
\end{eqnarray*}
and $\forall t$
\begin{eqnarray*}
\argmax_{1 \le i \le n} \delta_{t}^\prime(i) \;\;=\;\;
\argmax_{1 \le i \le n} \delta_{t}(i)
\end{eqnarray*}

The rationale behind this is to facilitate estimating the model parameters
from sparse data.
In more detail, it is easy to estimate $P({\it tag\/}\mid {\it word\/})$ for
a previously unseen word by backing off to statistics derived from words
that end with the same sequence of letters (or based on other surface cues),
whereas directly estimating $P({\it word\/}\mid {\it tag\/})$ is more 
difficult.
This is particularly useful for languages with a rich inflectional and
derivational morphology, but also for English: for example, the suffix
``-tion'' is a strong indicator that the word in question is a noun;
the suffix ``-able'' that it is an adjective.

More technically, the lexicon is organised as a reverse-suffix tree,
and smoothing the probability estimates is accomplished by blending the 
distribution at the current node of the tree with that of higher-level
nodes, corresponding to (shorter) suffixes of the current word (suffix).
The scheme also incorporates probability distributions for the set of
capitalized words, the set of all-caps words and the set of infrequent
words, all of which are used to improve the estimates for unknown words.
Employing a small amount of back-off smoothing also for the known words
is useful to reduce lexical tag omissions.
Empirically, looking two branching points up the tree for known words,
and all the way up to the root for unknown words, proved optimal.
The method for blending the distributions applies equally well to smoothing
the transition probabilities $p_{ij}$, i.e., the tag N-gram probabilities,
and both the
scheme and its application to these two tasks are described in detail in
\cite{Samuelsson:96}, where it was also shown to compare favourably to 
(deleted) interpolation, see \cite{Jelinek;Mercer:80}, even when the
back-off weights of the latter were optimal.

The $\delta$ variables enable finding the most probable state sequence
under the HMM, from which the most likely assignment of tags to words can
be directly established.
This is the normal modus operandi of an HMM decoder. 
Using the $\gamma$ variables, we can calculate the probability of being in
state $s_i$ at string position $t$, and thus having emitted $w_{k_t}$ from
this state, conditional on the entire word string.
By summing over all states that would assign the same tag to this word,
the individual probability of each tag being assigned to any particular
input word, conditional on the entire word string, can be calculated:
\begin{eqnarray*}
\lefteqn{P(X_t=x_i \mid \mbox{\bf W}) \;\;=}\\
&=&
\sum_{s_j : x_j = x_i} P(S_t=s_j \mid \mbox{\bf W}) \;\;=\;\;
\sum_{s_j : x_j = x_i} \gamma_t(j)
\end{eqnarray*}
This allows retaining multiple tags for each word by simply discarding
only low-probability tags; those whose probabilities are below some 
threshold value.
Of course, the most probable tag is never discarded, even if its probability
happens to be less than the threshold value.
By varying the threshold, we can perform a recall-precision,
or error-rate-ambiguity, tradeoff.
A similar strategy is adopted in \cite{deMarcken:90}.

\section{Experiments}

The statistical tagger was trained on 357,000 words from the Brown
corpus \cite{Francis:82}, reannotated using the EngCG annotation
scheme (see above).  In a first set of experiments, a 35,000 word
subset of this corpus was set aside and used to evaluate the tagger's
performance when trained on successively larger portions of the
remaining 322,000 words.  The learning curve, showing the error rate
after full disambiguation as a function of the amount of training data
used, see Figure~\ref{FigX}, has levelled off at 322,000 words,
indicating that little is to be gained from further training.  
We also note that the absolute value of the error rate is 3.51\%
--- a typical state-of-the-art figure.
Here, previously unseen words contribute 1.08\% to the total error
rate, while the contribution from lexical tag omissions is 0.08\%.
95\% confidence intervals for the error rates would range from
$\pm$ 0.30\% for 30,000 words to $\pm$ 0.20\% at 322,000 words.


\inspsfig{FigX}{Learning curve for the statistical tagger on the Brown corpus.}%
{295}{151}{-80}{-25}{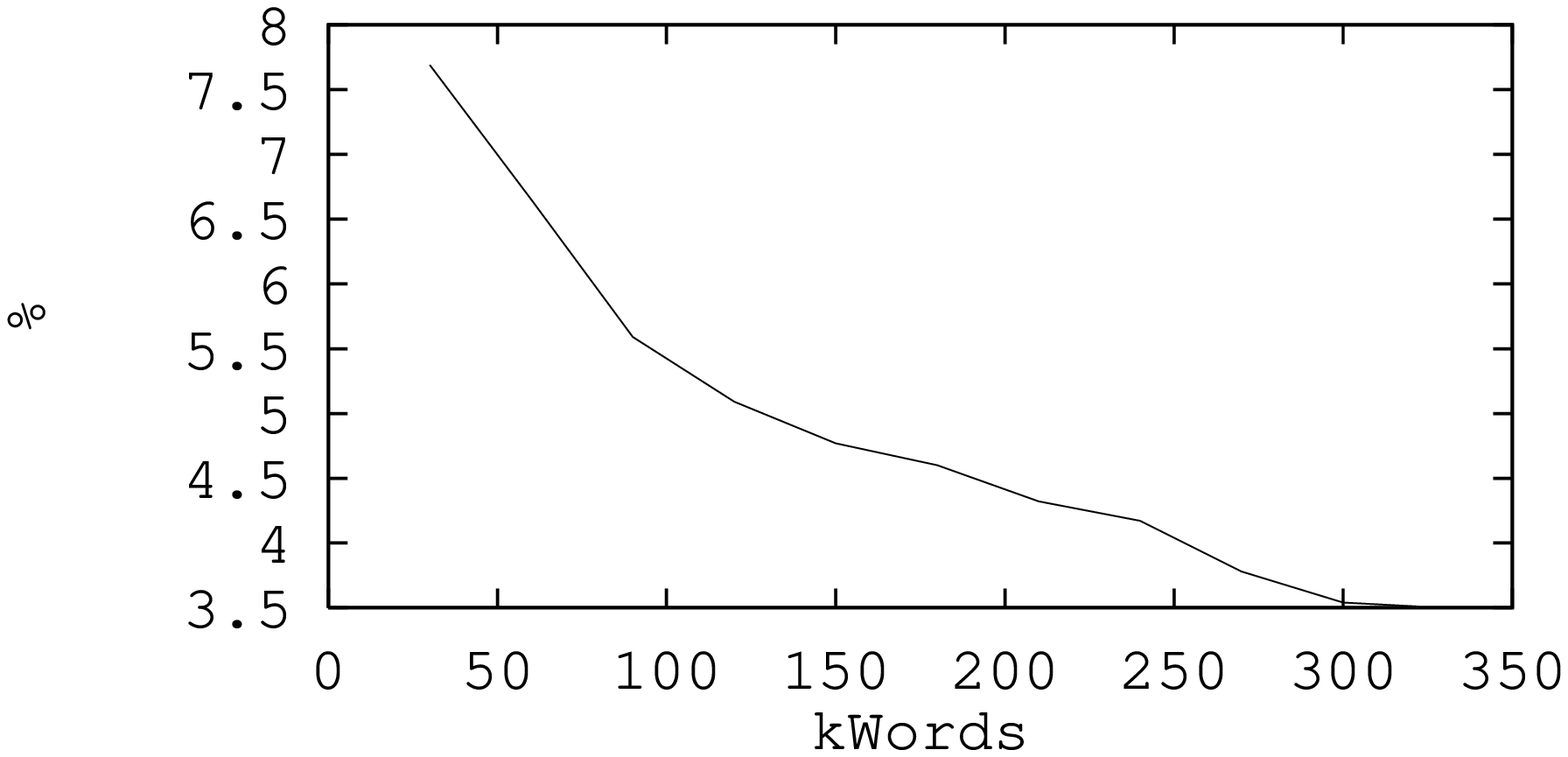}{50}{50}

The tagger was then trained on the entire set of 357,000 words and 
confronted with the separate 55,000-word benchmark corpus, and run both
in full and partial disambiguation mode.
Table~\ref{Table} shows the error rate as a function of remaining 
ambiguity (tags/word) both for the statistical tagger, and for the EngCG-2
tagger.
The error rate for full disambiguation using the $\delta$ variables
is 4.72\% and using the $\gamma$ variables is 4.68\%, both $\pm 0.18\%$
with confidence degree 95\%.
Note that the optimal tag sequence obtained using the $\gamma$ variables
need not equal the optimal tag sequence obtained using the $\delta$
variables.
In fact, the former sequence may be assigned zero probability by the
HMM, namely if one of its state transitions has zero probability.

\begin{table}
\begin{center}
\begin{tabular}{c|ccc}
Ambiguity       &\multicolumn{3}{c}{Error rate (\%)}\\
(Tags/word)     &\multicolumn{2}{c}{Statistical Tagger} &EngCG\\
                &($\delta$)&($\gamma$)& \\
\hline
1.000   &4.72   &4.68   & \\
1.012   &       &4.20   & \\
1.025   &       &3.75   & \\
1.026   &       &(3.72)        &0.43\\
1.035   &       &(3.48)        &0.29\\
1.038   &       &3.40   & \\
1.048   &       &(3.20)        &0.15\\
1.051   &       &3.14   & \\
1.059   &       &(2.99)        &0.12\\
1.065   &       &2.87   & \\
1.070   &       &(2.80)        &0.10\\
1.078   &       &2.69   & \\
1.093   &       &2.55   & \\
\hline
\end{tabular}
\end{center}
\caption{Error-rate-ambiguity tradeoff for both taggers on the benchmark corpus. 
  Parenthesized numbers are interpolated.}
\label{Table}
\end{table}

Previously unseen words account for 2.01\%, and lexical tag omissions
for 0.15\% of the total error rate. These two error sources
are together exactly 1.00\% higher on the benchmark corpus
than on the Brown corpus, and account for almost the entire difference
in error rate.  They stem from using less complete lexical
information sources, and are most likely the effect of a larger
vocabulary overlap between the test and training portions of the Brown
corpus than between the Brown and benchmark corpora.

The ratio between the error rates of the two taggers with the same
amount of remaining ambiguity ranges from 8.6 at 1.026 tags/word
to 28.0 at 1.070 tags/word.
The error rate of the statistical tagger can be further decreased, at the
price of increased remaining ambiguity, see Figure~\ref{FigY}.
In the limit of retaining all possible tags, the residual error rate
is entirely due to lexical tag omissions, i.e., it is 0.15\%, with in
average 14.24 tags per word.
The reason that this figure is so high is that the unknown words, which
comprise  10\% of the corpus, are assigned all possible tags as they are
backed off all the way to the root of the reverse-suffix tree.

\inspsfig{FigY}{Error-rate-ambiguity tradeoff for the statistical tagger
on the benchmark corpus.}%
{295}{151}{-80}{-25}{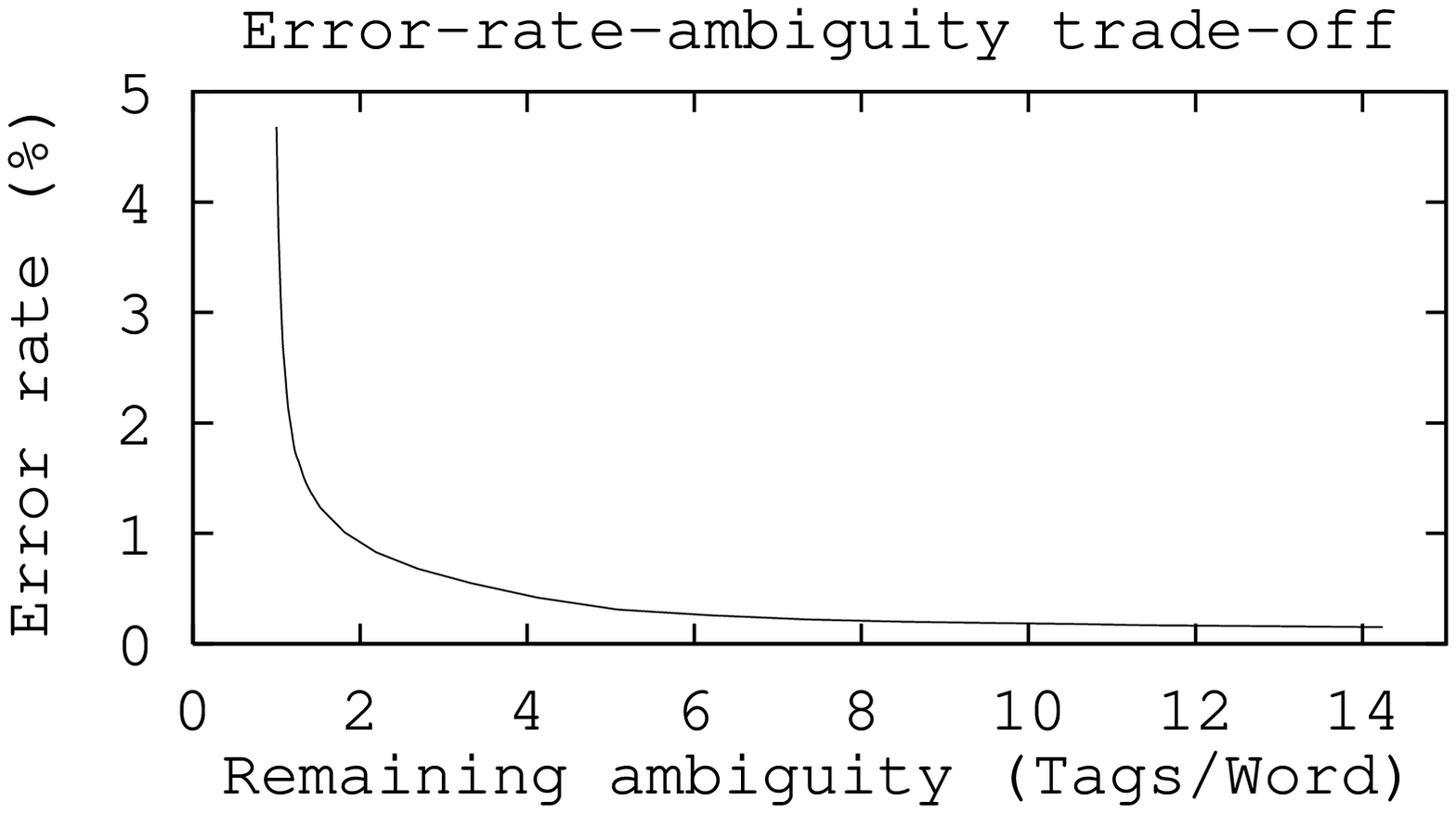}{50}{50}

\section{Discussion}

Recently voiced scepticisms concerning the superior EngCG tagging
results boil down to the following:

\begin{itemize}
\item The reported results are due to the simplicity of the tag set
  employed by the EngCG system.
\item The reported results are an effect of trading high ambiguity
  resolution for lower error rate.
\item The results are an effect of so-called priming of the human
  annotators when preparing the test corpora, compromising the
  integrity of the experimental evaluations.
\end{itemize}

In the current article, these points of criticism were investigated. A
state-of-the-art statistical tagger, capable of performing
error-rate-ambiguity tradeoff, was trained on a 357,000-word portion of
the Brown corpus reannotated with the EngCG tag set, and both taggers
were evaluated using a separate 55,000-word benchmark corpus new to both
systems.
This benchmark corpus was independently disambiguated by two
linguists, without access to the results of the automatic taggers.
The initial differences between the linguists' outputs (0.7\% of all
words) were jointly examined by the linguists; practically all of them
turned out to be clerical errors (rather than the product of genuine
difference of opinion).

In the experiments, the performance of the EngCG-2 tagger was radically
better than that of the statistical tagger: at ambiguity levels common
to both systems, the error rate of the statistical tagger was 8.6 to
28 times higher than that of EngCG-2.  We conclude that neither the tag
set used by EngCG-2, nor the error-rate-ambiguity tradeoff,
nor any priming effects can possibly explain the observed difference
in performance.

Instead we must conclude that the lexical and contextual information
sources at the disposal of the EngCG system are superior.
Investigating this empirically by granting the statistical tagger
access to the same information sources as those
available in the Constraint Grammar framework constitutes future work.

\section*{Acknowledgements}

Though Voutilainen is the main author of the EngCG-2 tagger, the
development of the system has benefited from several other
contributions too. Fred Karlsson proposed the Constraint Grammar
framework in the late 1980s. Juha Heikkil{\"a} and Timo J{\"a}rvinen
contributed with their work on English morphology and lexicon. Kimmo
Koskenniemi wrote the software for morphological analysis. Pasi
Tapanainen has written various implementations of the CG parser,
including the recent CG-2 parser \cite{PT:96}.

The quality of the investigation and presentation was boosted by a
number of suggestions to improvements and (often sceptical) comments
from numerous ACL reviewers and UPenn associates, in particular from
Mark Liberman.

\begin{table*}
{\small 
\begin{center}
\begin{tabular}{lll}
\multicolumn{3}{l}{\large\bf Appendix: Reduced EngCG tag set}\\
\\
\begin{tabular}{l}
\hline
Punctuation tags:\\
\hline
@colon\\
@comma\\
@dash\\
@dotdot\\
@dquote\\
@exclamation\\
@fullstop\\
@lparen\\
@rparen\\
@rparen\\
@rparen\\
@rparen\\
@lquote\\
@rquote\\
@slash\\
@newlines\\
@question\\
@semicolon\\
\hline
Word tags:\\
\hline
A-ABS\\
A-CMP\\
A-SUP\\
ABBR-GEN-SG/PL\\
ABBR-GEN-PL\\
ABBR-GEN-SG\\
ABBR-NOM-SG/PL\\
ABBR-NOM-PL\\
ABBR-NOM-SG\\
ADV-ABS\\
ADV-CMP\\
ADV-SUP\\
ADV-WH\\
BE-EN\\
\end{tabular}
&
\begin{tabular}{l}
BE-IMP\\
BE-INF\\
BE-ING\\
BE-PAST-BASE\\
BE-PAST-WAS\\
BE-PRES-AM\\
BE-PRES-ARE\\
BE-PRES-IS\\
BE-SUBJUNCTIVE\\
CC\\
CCX\\
CS\\
DET-SG/PL\\
DET-SG\\
DET-WH\\
DO-EN\\
DO-IMP\\
DO-INF\\
DO-ING\\
DO-PAST\\
DO-PRES-BASE\\
DO-PRES-SG3\\
DO-SUBJUNCTIVE\\
EN\\
HAVE-EN\\
HAVE-IMP\\
HAVE-INF\\
HAVE-ING\\
HAVE-PAST\\
HAVE-PRES-BASE\\
HAVE-PRES-SG3\\
HAVE-SUBJUNCTIVE\\
I\\
INFMARK\\
\end{tabular}
&
\begin{tabular}{l}
ING\\
N-GEN-SG/PL\\
N-GEN-PL\\
N-GEN-SG\\
N-NOM-SG/PL\\
N-NOM-PL\\
N-NOM-SG\\
NEG\\
NUM-CARD\\
NUM-FRA-PL\\
NUM-FRA-SG\\
NUM-ORD\\
PREP\\
PRON\\
PRON-ACC\\
PRON-CMP\\
PRON-DEM-PL\\
PRON-DEM-SG\\
PRON-GEN\\
PRON-INTERR\\
PRON-NOM-SG/PL\\
PRON-NOM-PL\\
PRON-NOM-SG\\
PRON-REL\\
PRON-SUP\\
PRON-WH\\
V-AUXMOD\\
V-IMP\\
V-INF\\
V-PAST\\
V-PRES-BASE\\
V-PRES-SG1\\
V-PRES-SG2\\
V-PRES-SG3\\
V-SUBJUNCTIVE\\
\end{tabular}
\end{tabular}
\end{center}
} 
\end{table*}

\end{document}